\documentclass[12pt]{paper}
\usepackage[
  margin=3.55cm,
  includefoot,
  footskip=30pt,
]{geometry}
\usepackage{layout}                                                 
\usepackage{graphicx}
\usepackage{amsmath}
\usepackage{amssymb}
\usepackage{amstext}
\usepackage{amsthm}
\usepackage{amssymb}
\usepackage{times}

\usepackage{todonotes}

\usepackage{rotating}

\usepackage[normalem]{ulem}

\usepackage{multirow}

\usepackage{caption}

\usepackage[ruled,vlined]{algorithm2e}

\usepackage{verbatim}

\usepackage{xcolor}
\usepackage{etoolbox}
\patchcmd{\thebibliography}
  {\settowidth}
  {\setlength{\parsep}{0.25pt}\setlength{\itemsep}{0pt plus 0.52pt}\settowidth}
  {}{}
\apptocmd{\thebibliography}
  {\small}
  {}{}
\newtheorem{problem}{Problem}

\usepackage{lipsum}
\usepackage{hyperref}
\hypersetup{
    colorlinks=true,
    linkcolor=blue,
    filecolor=magenta,
    urlcolor=cyan,
}
\DeclareMathOperator*{\argmin}{argmin}

\author{{\small{Stefano Coniglio, Anthony J. Dunn, and Alain B. Zemkoho}}\\
\emph{\small{School of Mathematical Sciences, University of Southampton, United Kingdom}}\\
\small{$\{$\texttt{s.coniglio}, \texttt{a.j.dunn}, \texttt{a.b.zemkoho}$\}$\texttt{@soton.ac.uk}}}

\numberwithin{equation}{section}
\begin{document}

\title{Infrequent adverse event prediction in low-carbon energy production using machine learning}
\maketitle


\begin{abstract}
  We address the problem of predicting the occurrence of infrequent adverse events in the context of predictive maintenance.
  We cast the corresponding machine learning task as an imbalanced classification problem and propose a framework for solving it that is capable of leveraging different classifiers in order to predict the occurrence of an adverse event before it takes place.
  In particular, we focus on two applications arising in low-carbon energy production: foam formation in anaerobic digestion and condenser tube leakage in the steam turbines of a nuclear power station.
  The results of an extensive set of computational experiments show the effectiveness of the techniques that we propose.
\end{abstract}


\section{Introduction}\label{Intro}

In this paper, we address the predictive-maintenance problem of predicting the occurrence of {\em infrequent adverse events} with machine learning techniques.
The method we propose takes operating conditions as input and provides as output an indication of whether the adverse event is likely to occur within a given (appropriate) time frame.
From an engineering perspective, our work falls within the remit of \emph{predictive maintenance}, which, traditionally, encompasses the design of condition-driven preventative maintenance programmes~\cite{mobley2002introduction}.
For more references to the area, we refer the reader to~\cite{omshi2020dynamic,barlow2020performance,crowder2007scheme,van2020effect,de2019review}.
From a machine-learning perspective (see~\cite{gambella2020optimization} for a recent survey on the optimisation aspects of machine learning), the problem we tackle can be viewed as a supervised anomaly-detection task~\cite{chandola2009anomaly}.

Throughout the paper, we focus on two applications arising in low-carbon (bio and nuclear) energy production: foam formation in anaerobic digestion and condenser tube leakage in the steam turbines of a nuclear power station.
Further details on the two applications are given in the next subsections.

This work is done in collaboration with the energy team of Decision Analysis Services Ltd (DAS Ltd), a consultancy group based in Bristol, UK, with expertise in various areas, including engineering and energy.

\subsection{Foaming in anaerobic digestion for biogas production}\label{anaerobic digestion}

The first application we consider arises in the context of bioenergy for the production of biogas via {\em Anaerobic Digestion} (AD).

In a typical AD process, sludge is fed into a digester, where it is broken down into micro organisms to release biogas which is then collected from the top of the digester and burned to produce energy.
This feed sludge is comprised of organic matter such as food waste, agricultural waste, or crop feed.
Under normal operations, gas bubbles rise from feed sludge as it is digested, and then collapse releasing biogas.
This biogas is then burned in order to power a turbine so to produce electricity.
Under certain conditions, the gas bubbles may take longer to collapse than it takes for new bubbles to form, resulting in the creation of \emph{foam}~\cite{ganidi2009anaerobic}.

Foaming is often a major issue in AD, as it can block the gas outlet, resulting in the digester having to be shut down for cleaning operations to take place.
As cleaning can take days and, during this time, the digester cannot be used for biogas production, foam formation can have a considerable impact on the energy output.
In some extreme cases, it can even lead to the roof of the digester being blown off due to the pressure building up due to the gas outlet and the pressure release valve becoming blocked.
Foaming can be treated using an {\em anti-foaming agent} which, upon introduction into the digester, causes the bubbles to collapse.
This, however, requires the foaming phenomenon to be detected early enough to prevent it from causing serious problems.

For this application, our aim is to create a predictive maintenance tool capable of reliably predicting foaming with enough warning for the plant operators to administer the anti-foaming agent and subdue the foaming phenomenon before it can damage the digester.

While foaming in AD is a well-researched area, see, for instance,~\cite{ganidi2009anaerobic}, most of the works, such as, e.g.,~\cite{kanu2015understanding, kanu2018biological}, analyse the phenomenon from a chemical or engineering perspective and, to the best of our knowledge, only a few attempts have been made at modelling it using machine learning methods.
See, e.g.,~\cite{dalmau2010selecting, Fernandes2014ANNforAD}.
In particular, most data driven approaches rely on the design of a knowledge-based system (KBS) and, thus, require in-depth knowledge of the specific digester in question. See, e.g.,~\cite{dalmau2010model, gaida2012state, kanu2018biological}.
Often, they also rely on a deep knowledge of the chemical composition of the feed stock, also known as {\em feed sludge characteristics}, as well on time-series sensor readings of the conditions within the digester, typically referred to as {\em operating characteristics}.
While it has been shown that relying on feed sludge characteristics can be more effective than simply considering operating characteristics, monitoring them is not always feasible in many industrial applications (although they can be cheaply and easily collected).
In spite of this, in the context of foaming-formation in AD, machine learning models have only been used for a limited extent, such as for state estimation using, e.g., neural networks to predict methane production~\cite{gaida2012state, Fernandes2014ANNforAD}, and for gaining insights into which variables could be the best indicators of foaming~\cite{dalmau2010selecting}.
Crucially, such approaches are often unable to predict foaming before it occurs, as, typically, they provide a warning only when the adverse event is already in a developed stage and it is too late for taking any mitigating actions~\cite{dalmau2010model}.

On the contrary, the methods we propose in this paper allow for predicting foaming formation within a user-specified warning window from operating characteristics alone.
As the injection of the anti-foaming agent into the digester appears to be the most popular approach to prevent foam from building up, thanks to predicting foaming sufficiently ahead of its occurrence our method can result in a significant reduction in the amount of anti-foaming agent required to avoid foaming, thus leading to a considerable cost reduction.

\subsection{Condenser tube leaks in nuclear power production}\label{Condenser Fouling}

Te second application we consider arises in the context of {\em nuclear power production} (NPP).
In particular, we focus on the operation of steam turbines converting thermal energy from steam to electrical power.
Under normal operations, steam is generated in an array of boilers and passed through the turbine rotors causing them to rotate, thus generating power---a turbine is said to be \emph{on-load} if it is rotating and hence producing power, and \emph{off-load} if it is not.
The steam is then passed through a condenser, where it is cooled down to liquid water and recirculated.
The condenser consists of thousands of titanium tubes containing sea water, which acts as the primary coolant.

Under normal operating conditions, the sea water circuit remains isolated from the steam circuit.
It is, however, possible for a leak to form in one of the condenser tubes, causing sea water to contaminate the steam circuit.
Such a leak is typically caused by the formation of deposits on heat-transfer surfaces (\emph{fouling})---for more details on condenser fouling within nuclear power production, we refer the reader to~\cite{muller1999cooling}.

In the event of a leak in one of the tubes of the condenser, the steam turbine is automatically tripped, and remedial actions are taken to fix the leak off-load.
Besides maintenance costs, an unplanned trip causes a loss of revenue proportional to the time needed to fix the leak and restart the power plant, which can be quite substantial.

If warning of a tube leak were provided prior to its advent, the power output could be reduced and the turbine in question isolated and repaired on-load before the plant is brought back to full power, resulting in much smaller generation losses.
Through communication with DAS Ltd, we estimate that such a predictive functionality could yield saving up to \pounds350,000 per day of warning provided and per tube leak.
To the best of our knowledge, no machine learning techniques have been developed so far to achieve such a goal, which we set out to achieve with the work in this paper.

\subsection{Contributions and outline of the paper}

\begin{figure}[!ht]
  \centering
  \includegraphics[width=\textwidth]{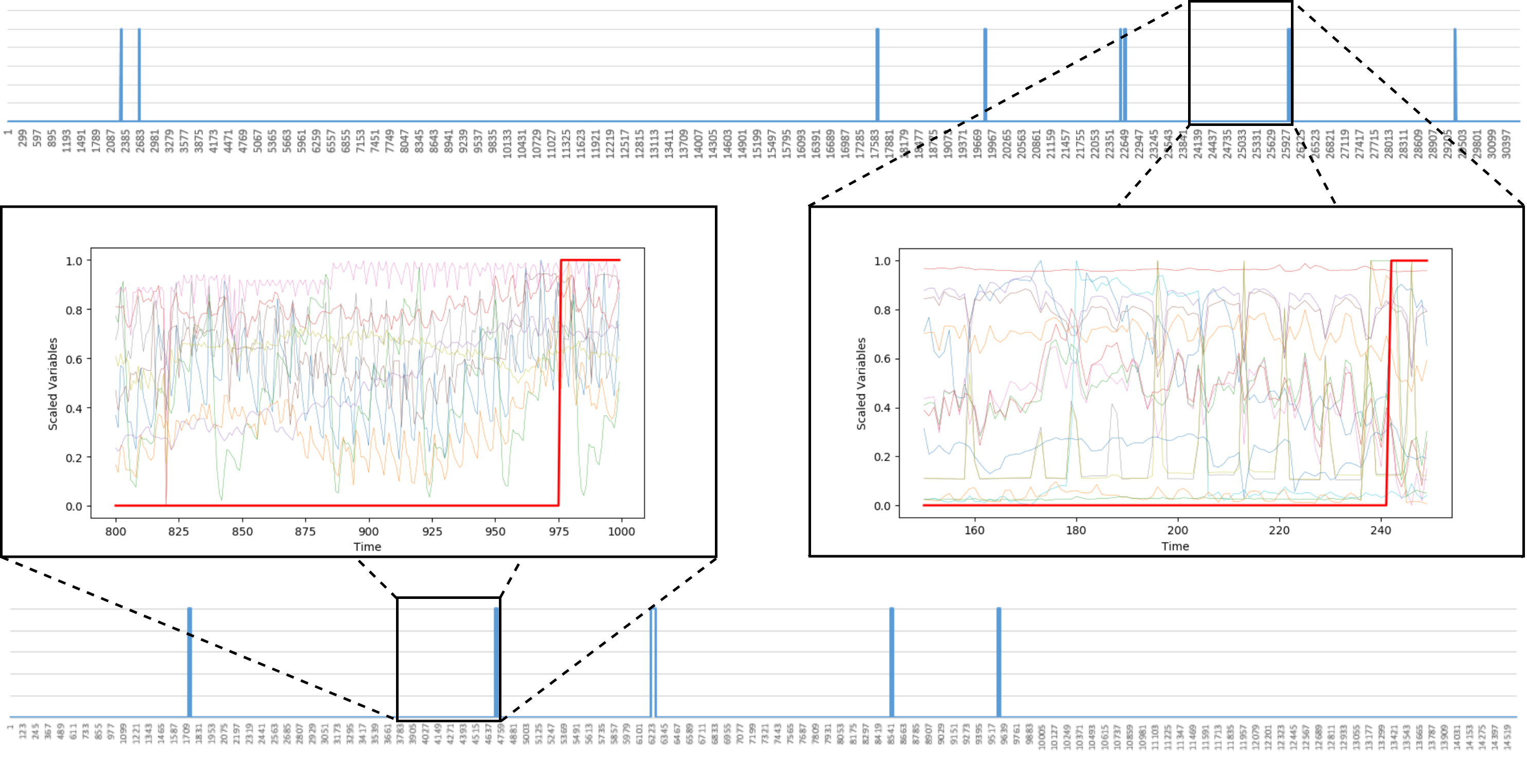}
  \caption{Scaled condenser variables (top) and anaerobic digester variables (bottom).}\label{Time Line}
\end{figure}

The structure of the two applications described above, AD and NPP, is strikingly similar.
For each of them, we have an adverse event occurring at, unpredictable at present, time intervals and a large amount of time-series data consisting of regular readings of operating conditions.
Figure~\ref{Time Line} better illustrates this similarity, reporting two time series provided to us by DAS Ltd.
The top one shows adverse events of condenser tube leaks occurred from 2009 to 2019 in an NPP plant.
The bottom one reports adverse events of foaming that occurred between December 2015 and July 2017.
The figure zooms in on two sections of the data, showing the behaviour of the variables in the run-up to an event.
As one can see, the plots do not exhibit a clear trend or a clear pattern leading to the occurrence of an event, indicating that the prediction task as hand is a non-trivial one.

Our main goal in this paper is to propose a method for
leveraging
various machine learning techniques in order to capture the hidden phenomena leading to adverse events in both applications sufficiently ahead of time to allow for effective remedial actions to be taken.
%
As we will show, the proposed method can be used to assess and fine tune virtually any classification algorithm (in particular, in the experiments we report we use this method to evaluate 11 popular classification algorithms).

The paper is organised as follows.
Section~\ref{Formalising The Classification Problem} outlines the general framework that we introduce to build the predictive maintenance model.
Section~\ref{Solving the problem} introduces the algorithm that we propose to solve the resulting learning problem, as well as a set of appropriate modifications that are carried out on top of the basic algorithm to achieve a higher efficiency in our specific context.
Section~\ref{Experiments} assesses the quality of our method on the two applications we consider via extensive computational experiments in which, in particular, difference machine-learning paradigms are compared and contrasted.
It also indicates which predictive maintenance models are advisable for use on each of them, as well as which machine learning techniques achieve, in practice, the best performance.
Section~\ref{Conclusion} concludes the paper by drawing concluding remarks.


\section{Classification model}\label{Formalising The Classification Problem}

Given a time horizon $T = \{1, \dots, |T|\}$, let $(x_t, \, y_t)_{t\in T}$ be a time series where, for each $t \in T$, each data point $x_t \in \mathbb{R}^d$ consists of an observation of $d$ features at time $t$ and $y_t$ is a binary label denoting whether the adverse event occurs at time $t$ or not.
More formally, the couple $(x_t, \, y_t)_{t\in T}$ is such that:
\begin{equation}\label{basic data}
  x_t \in \mathbb{R}^d \;\, \mbox{ and }
  \;\, y_t = \left\{\begin{array}{ll}
                      1 & \mbox{ if the event occurs at time } t,\\
                      0 & \mbox{ otherwise.}
                    \end{array}\right.
                \end{equation}
                %


Adopting a common pre-processing technique often used in time-series forecasting and referred to as {\em leading} or {\em lag inclusion}, for each $t \in T$ we introduce a matrix $\tilde{X}_t \in \mathbb{R}^{d \times (\tau+1)}$ defined as follows:
\begin{equation}\label{pattern}
 \tilde{X}_t := (x_{t-\tau}, \ldots, \; x_t),
\end{equation}
where $\tau$ is the number of lags that we wish to include.
For each $t \in T$, we refer to $\tilde{X}_t$ as a \textit{pattern}.

The problem of adverse event prediction from time-series data which we tackle in the paper can now be formalised as the problem of determining a function $h$ which, for a given pattern $\tilde{X}_t$, returns the associated label $y_t$ with minimal error.
This problem can be formally stated as follows:
\begin{problem}\label{intuitive_problem}
  Given $\left(\tilde{X}_t,\, y_t\right)_{t\in T}$, find a function $h: \mathbb{R}^{d\times(\tau+1)}\rightarrow \{0,1\}$ which minimises a loss function $\mathcal{L}: \mathbb{R}^{|T|} \rightarrow \mathbb{R}$ of the prediction errors $|y_t - h(\tilde{X}_t)|$, $t \in T$.
\end{problem}
\noindent The problem calls for a function $h \, : \tilde{X}_t \mapsto \{0, 1\}$ (also commonly referred to as a rule, a predictor, a hypothesis, or a classifier~\cite{shalev2014understanding}) which best approximates the true (unknown) labelling function $f \, : \tilde{X}_t \mapsto \tilde y_t, t \in T$, which maps each data point to the corresponding label, by minimising a {\em loss function} $\mathcal{L}$.
A wide range of loss functions have been developed in the literature, each suitable for a specific problem class---for more details, we refer to reader to~\cite{Vapnik1995}.

As it is not difficult to see, an algorithm for solving Problem~\ref{intuitive_problem} would result in a tool only capable of predicting the event when it is already occurring.
Such a tool would be of little use in our context as, once an adverse event is occurring, it is typically too late to prevent it or even to minimise the damage that it will cause.
Our aim is, therefore, to predict the adverse event before it actually occurs with enough warning so that maintenance can be carried out to prevent or minimise the damage before it is too late.

Let $\omega$ be the minimum length of the warning window that is required for the application at hand.
The value of $\omega$ should be carefully selected to be large enough to allow sufficient time for the required maintenance actions to take place and small enough to be reasonably close to the event (as too large an $\omega$ would lead to $\tilde y_t = 1$ for almost all $t \in T$, rendering the prediction, {\em de facto}, useless).

To be able to capture an adverse event at least $\omega$ steps before its occurrence, we introduce a new set of labels $\tilde{y}_t \in \{0,1\}$, $t \in T$, defined in such a way that, for a given $\omega \in \mathbb{N}$, the following holds:
\begin{equation}\label{association_label}
  \tilde{y}_t = 1 \quad \Leftrightarrow \quad \exists t^\prime \in \left\{t,\, \dots, \, t+\omega-1\right\}: \;\, y_{t^\prime} = 1.
\end{equation}
We refer to such labels as {\em warning labels}.
According to this definition, $\tilde{y}_t$ is equal to~$1$ if and only if an adverse event occurs in the time window starting at time $t$ and ending at time $t+\omega-1$, whereas it is equal to $0$ otherwise.

In order to predict an adverse event \emph{before it occurs} and in light of the definition of $\tilde y_t$, we introduce the following learning problem:
\begin{problem}\label{problem}
  Given $\left(\tilde{X}_t,\, \tilde{y}_t\right)_{t\in T}$, find a function $h: \mathbb{R}^{d\times(\tau+1)}\rightarrow \{0,1\}$ which minimises a loss function $\mathcal{L}: \mathbb{R}^{|T|} \rightarrow \mathbb{R}$ of the prediction errors $|\tilde y_t - h(\tilde{X}_t)|$.
\end{problem}
With Problem~\ref{problem}, we aim to find a function $h$ which, given a pattern~$\tilde{X}_t$,
can recover the pattern's true warning label~$\tilde{y}_t$, thereby indicating if an event will occur within the following $\omega$ time intervals so to allow for remedial actions to be taken in time.

Problem~\ref{problem} entails a binary classification task.
For a chosen classifier $C$, solving Problem~\ref{problem} corresponds to solving the following optimisation problem, where $H_C$ is the family of prediction rules that correspond to classifier $C$ (e.g., the set of linear discriminations in linear-kernel SVM) and $\mathcal{L}_C$ is the loss function associated with the training of that classifier:
\begin{problem}\label{prob:3}
  Given $\left(\tilde{X}_t,\, \tilde{y}_t\right)_{t\in T}$, find $h^*$ such that:
\begin{equation*}
 h^* \, \in \argmin_{h \in H_C} \mathcal{L}_{C} \left(\big(h(\tilde{X}_1),\tilde{y}_1\big), \dots, \big(h(\tilde{X}_{|T|}),\tilde{y}_{|T|}\big)\right).
\end{equation*}
\end{problem}
Solving Problem~\ref{prob:3} for a given classifier is just a step of the method we propose in the next section.
In our method, Problem~\ref{prob:3} can be solved by adopting, virtually, any classifier (we will consider 11 such classifiers in Section~\ref{Experiments}).
In the next section, we outline the main challenges that one faces when tackling the problem with an off-the-shelf method, and the techniques that we propose to avoid them.


\section{Solving the classification problem}\label{Solving the problem}

We adopt the customary method of constructing a prediction rule by fitting a classifier to a {\em training set} and then testing such a rule on a reserved, unseen {\em testing set}.
The classifier which performs best at predicting the correct warning labels for the testing set is considered to be the one which best approximates the true labelling function.
We refer to this procedure as {\em model-selection}.

As it is typical when assessing classification accuracy, we repeat our training and testing via {\em $k$-fold cross validation} (CV).
In CV, the data set is divided into a set of $k$ roughly equal-sized subsets (or {\em folds}) by, typically, randomly sampling without replacement.
At each of $k$ iterations, the $i$-th fold, with $i = 1, \dots, k$, is selected for testing, a classifier is trained on the remaining $k-1$ folds, and its performance is measured on the testing fold.
This results in $k$ measures of the accuracy of the classification algorithm which are then averaged into a single, final measure.

{\em Balanced accuracy} is the accuracy metric we consider in this paper.
Such a metric is defined as the average of two commonly used metrics, {\em sensitivity} (equal to the ratio between the number of true positives and the total number of positives) and {\em specificity} (equal to the ratio between the number of true negatives and the total number of negatives).\footnote{Other options are possible.
  For instance, the {\em area under an ROC} (receiver operating characteristic) {\em curve} is a very popular metric for comparing classification algorithms on imbalanced data sets~\cite{brown2012experimental}.
  We remark that a large area under an ROC curve only indicates that there exists a range of decision thresholds which have good corresponding true-positive and false-positive rates.
  To use this method in practice, we would need to select one of these decision thresholds with optimal true-positive and false-positive rates, which would require an additional round of CV.
  As, in our approach, the number of folds we can separate our data into is limited by the number of events we have and the latter is fairly small, performing this additional round of CV is not practical.}

Due to the nature of the learning tasks we tackle, a number of crucial issues are likely to arise when resorting to this classical method for training and testing the classification algorithm we propose.
We address each of such issues in the following, presenting the appropriate techniques we adopt for facing it.

\subsection{Drawbacks to standard approaches}

We begin by establishing the need for chronology.
It is common to avoid shuffling time series data.
This is especially true when creating patterns via lag inclusion as, by randomly sampling patterns into training and testing sets, it is likely for a given pattern $\tilde{X}_t$ in the testing set that either $\tilde{X}_{t-1}$ or $\tilde{X}_{t+1}$ be contained in the training set.
As the features that are observed on both the AD and the NPP applications do not vary drastically over a single time interval, it is likely for the pattern $\tilde{X}_t$ to be very similar to both $\tilde{X}_{t-1}$ and $\tilde{X}_{t+1}$.
This would result in the testing set containing patterns which were
``almost seen'' in the training phase, resulting in solutions which, while exhibiting a very high predictive accuracy on the testing set, are likely to enjoy poor generalisation capabilities when trialled on new, unseen data such as those that would be used when adopting the learnt prediction rule as a predictive-maintenance tool.

In light of this, an intuitive fold sampling method would be to simply partition the data into equal sized folds, preserving chronology in each of them.
There are, however, significant issues with the application of this method to our problem, which we now outline.

Due to the infrequent occurrence of adverse events, the data sets of the two applications we consider are likely to suffer from a large \emph{class imbalance}, i.e, they are likely to contain a far greater number of patterns which are not associated to the event than those which are.
Class imbalance is known to be a problem for most classification algorithms, as their solutions are likely to overfit the majority class.
In our case, this would result in predicting the non-occurrence of the adverse event in almost all the cases, rendering the prediction {\em de facto} useless from a predictive-maintenance perspective.

As the adverse events occur infrequently and irregularly in the two applications we consider, it is crucial to partition the data set into the $k$ folds in such a way that at least one adverse event occurs in each of them.
Failing to guarantee this would lead to at least a round of testing on a testing set that does not contain any adverse events.
Since any classifier which predicts (accurately or not) the non-occurrence of the adverse event would score extremely well in any such testing set regardless of how well it predicts the event itself, this would lead to a further bias towards the majority class.

Our binary warning labels simplify the state of the system (either the AD plant or the NPP turbine) to two states: the system being at risk of the adverse event occurring within the next $\omega$ time intervals if $\tilde y_t = 1$ and the system being in normal operations if $\tilde y_t = 0$.
In the time period immediately following an event, we would not expect the system to have returned to normal operations, typically due to changes caused either directly by the event or indirectly by the maintenance undertaken.
For this reason, while we would not label patterns corresponding to a point in time immediately following an event $\tilde y_t = 1$, they are not indicators that the system is in a stable state and so should not be labelled $\tilde y_t = 0$ either.
For this reason, retaining such points in our data set with either label may negatively affect the performance.
\subsection{Block sampling}

In this section, we propose a sub-sampling method for addressing all of the previously described drawbacks to fold sampling by partition.
We identify a set of $k$ non-overlapping and contiguous {\em blocks} of equal size $\beta$ within the data set, where $k$ is the number of events that occur in the entire data set.
We use $k$ here as we will later use these blocks as folds in $k$-fold CV.
Each block $T_i$, for $i \in \{1, \dots, k\}$, consists of the end of a warning event (i.e., the last time step in a sequence with $\tilde{y}_t = 1)$) together with the preceding $\beta$ time intervals.
Namely, we have a block $T_i$ for each $i = 1, \dots, k$ where:
\begin{equation*}
\begin{aligned}
  T_i := \{t-\beta, \ldots, t\}, \qquad \text{ where } \tilde{y}_t = 1 \wedge \tilde{y}_{t+1} = 0.
\end{aligned}
\end{equation*}

Due to the adoption of blocks, the data set used for solving the classification problem is partitioned as follows:
\begin{equation*}
  \left\{\left(\tilde{X}_t,\, \tilde{y}_t\right)_{t\in T_{1}}, \ldots, \left(\tilde{X}_t,\, \tilde{y}_t\right)_{t\in T_{k}}\right\}.
\end{equation*}
Given the $k$ blocks, we generate each fold by sampling, rather than individual patterns, an entire block.
The adoption of blocks naturally leads to sub-sampling the data into ($k$) folds without shuffling, thanks to which the folds are structured chronologically in such a way that each of them contains patterns that are {\em contiguous} in time.
This avoids the issue of shared visibility between training and testing sets that we mentioned before.

This method ensures than each fold contains exactly one event, removes all data from the unstable period which immediately follows events, and helps to address the class imbalance by dropping a large amount of data which is not related to the events.
In Section \ref{Experiments}, we will compare the use of these blocks as folds in our $k$-fold CV experiments to a simpler method where the entire data set is partitioned into equal sized folds within each of which chronology is preserved.

As a further technique to help mitigate the majority-class bias, we resort to under or oversampling when constructing the training sets from our folds (while retaining un-sampled data in the testing sets).
When undersampling, we sample, with replacement, a set from the majority class (in which the event does not occur) of the same cardinality as the minority class (in which the event does occur).
When oversampling, we sample, with replacement, a set from the minority class that has the same cardinality as the majority class.
Experiments on the befit of adopting such a technique will be highlighted in the next section.

\subsection{Event inclusion and exclusion}

We call all patterns whose original label $y_t$ is $1$ \textit{event patterns}.
As our aim is to build models capable of predicting if an event is likely to occur in the $\omega$ time intervals preceding it, we are not interested in accurately predicting if the event is currently occurring.
For this reason, we always remove all the event patterns from every CV fold when we use them for testing.

Our aim in this paper is to build models capable of predicting if an event is likely to occur in the next $\omega$ time intervals.
In particular, we are not interested in whether our models can accurately predict if an event is currently occurring.
For this reason, we remove all patterns whose original label $y_t$ is $1$ from our CV folds when we use them for testing.
We call such patterns \textit{event-patterns}.
We define a fold whose event patterns have been removed as:
\begin{equation}\label{BlockDefineNoEvent}
\begin{aligned}
  &\left(\tilde{X}_t,\, \tilde{y}_t\right)_{t\in T_{i}^\prime} && \text{with} &&&T_{i}^\prime = \left\{\left. t \in T_i  \right|\; y_t =0\right\}.
\end{aligned}
\end{equation}
Although we always exclude the event-patterns from our testing fold, we may choose to include or exclude the event-patters from our training folds. Classifiers are tested on $\{T_1^\prime,\, \ldots,\, T_k^\prime\}$, but can be trained on either $\{T_1,\, \ldots,\, T_k\}$ or $\{T_1^\prime,\, \ldots,\,  T_k^\prime\}$. We generally define a fold as
\begin{equation}\label{includeEvent}
\begin{aligned}
    &\left(\tilde{X}_t,\, \tilde{y}_t\right)_{t\in F_{i}} &&\text{with} &&&F_{i} = \begin{cases}T_i &\text{if } \;\;\; \text{include event}\; = \;\mathbf{true},\\T_i^\prime &\text{if } \;\;\; \text{include event}\; = \;\mathbf{false}.
    \end{cases}
\end{aligned}
\end{equation}
We define $\mathcal{F}$ as the union of the $k$ folds $F_1,\, \ldots,\, F_k$.

We may choose to include such event patterns in the training folds.
Doing so makes the implicit assumption that the behaviour of the system during the event itself is indicative of its behaviour leading up to the event.
Including these patterns would increase the number of patterns associated to the event, helping to reduce the large class imbalance and improving the strength of the classifiers.
On the contrary, excluding the event patterns makes the assumption that the behaviour of the system during the event be independent from its behaviour leading up to the event. If this is the case, including these patters would only add noise to the minority class, ultimately resulting in weaker classifiers.
For this reason, deciding whether to include or exclude the event-patterns from the training sets is case specific.
Preliminary experiments or in-depth knowledge of the system and of the event is required to make this decision.

This need to sometimes include the event-patterns in training but not in testing at each iteration of CV requires us to modify our $k$-fold CV method.
We define the CV accuracy for a given classifier as:
\begin{equation}\label{crossValidationAccuracy}
  \frac{1}{k}\sum_{i=1}^{k} \text{accuracy} \left(h_{\mathcal{F}\backslash F_i}(\tilde{X}_{T_{i}^\prime}), \; \tilde{y}_{T_i^\prime}\right),
\end{equation}
using our modified $k$-fold CV method.
In this formula, $\tilde{X}_{T_{i}^\prime}$ and $\tilde{y}_{T_i^\prime}$ are defined as described in~\eqref{BlockDefineNoEvent} with $h_{\mathcal{F}\backslash F_i}$, which corresponds to the decision rule of a classifier which has been trained on the data and labels with indices $\mathcal{F}\backslash F_k$, as given in~\eqref{includeEvent}.
As for the term ``\emph{accuracy}'', it represents an appropriate accuracy function such as balanced accuracy or the area under an ROC curve.

\subsection{Hyperparameter selection}

Modern classification algorithms often depend on one or more {\em hyperparameters}, whose tuning is crucial to obtain a high prediction accuracy.
Among different options, {\em grid search} (which consists in solving a learning problem for each discretised combination of values the hyperparameters can take) is, arguably, the most widely applied technique.

Hyperparameter tuning would normally be done at each iteration of the CV method.
After the testing fold is chosen, we would iteratively split the training set into a {\em validation set} and a new, smaller training set.
We would then perform grid search (or another method, such as random search) to find suitable values for the hyperparameters which result in the best performance on the validation set.

At each iteration of our $k$-fold CV for classification-algorithm evaluation, we will perform a ($k$-1)-fold CV in order to tune the hyperparameters.
The folds used in this ($k$-1)-fold CV will be the same $k$-1 folds originally assigned for training.
This means that all of the desirable features of our folds, such as chronology and ensuring that each fold contain exactly one event, will be preserved in the ($k$-1)-fold CV.
As with the basic $k$-fold method, we will have to modify this method to allow for including event patterns during training while excluding them during testing.

Let $\Lambda$ be the set of all combinations of values for the hyperparameter(s) that are being tested.
As the decision rule found by a classifier is dependant on its hyperparameter selection, we now redefine the decision rule as $h_{\lambda, A}$ where $\lambda$ is the chosen value of the hyperparameters and $A$ is the index vector of the data on which the classifier is trained.
Let $\lambda^{\mathcal{F}\backslash F_i}$ be the set of hyperparameter values which gives the best performance on the data with index set $\mathcal{F}\backslash F_i$.
Similarly, let $\alpha^{\mathcal{F}\backslash F_i}$ be the average CV accuracy of a classifier trained with the best hyperparameter selection.
Algorithm~\ref{Gridsearch} shows the modified version of ($k$-1)-fold CV for hyperparameter selection nested within $k$-fold CV for model selection.

\begin{algorithm}[h!]
\SetAlgoLined

 \KwResult{Average $k$-fold CV classification accuracy with tuned hyperparameters}
 \For{$F_i \in \mathcal{F} := \{F_1 \ldots F_k\}$}{
    $\alpha^{\mathcal{F}\backslash F_i} = 0$\\
    \For{$\lambda \in \Lambda$}{
        $\alpha \leftarrow \frac{1}{k-1}\sum_{F_j \in \mathcal{F}\backslash F_i} \text{accuracy}\left( h_{\lambda, (\mathcal{F}\backslash F_i)\backslash F_j}(\tilde{X}_{T_j^\prime}), \tilde{y}_{T_j^\prime} \right)$\\
        \If{$\alpha \geq \alpha^{\mathcal{F}\backslash F_i}$}{
            $\alpha^{\mathcal{F}\backslash F_i} = \alpha$\\
            $\lambda^{\mathcal{F}\backslash F_i} = \lambda$\:
        }
    }
 }
 return: $\frac{1}{k}\sum_{F_i \in \mathcal{F}}\text{accuracy} \left(h_{\lambda^{\mathcal{F}\backslash F_i}, \mathcal{F}\backslash F_i}(\tilde{X}_{T_i^\prime}), \tilde{y}_{T_i^\prime}\right)$
 \caption{Nested $k$-fold CV for model selection using gridsearch for hyperparameter selection}
 \label{Gridsearch}
\end{algorithm}

We note that no changes to the under and oversampling techniques designed to address class imbalance in the training of our classifiers are needed,  as these only apply to the training sets and not to the testing sets.

\section{Numerical experiments}\label{Experiments}

In this section, we assess the predictive power of the prediction method we proposed by experimenting on a data set corresponding to the two applications we focussed on, i.e., foaming formation in AD and condenser fouling in NPP.

\subsection{Setup}

For foaming formation in AD, we rely on a data set provided to us by DAS Ltd in collaboration with Andigestion Ltd, a UK-based AD company producing renewable energy for the national grid through the combustion of biogas obtained from natural waste.
The AD data we use consists in a time series collected from sensors at Andigestion Ltd's plant based in Holsworthy, UK.
The data set consists of 14,617 hourly readings of 9 numeric variables.
This amounts to 20 months of runtime from December 1st 2015 to July 31st 2017.
Over this period, 5 distinct foaming events occurred.

For condenser fouling in NPP, we consider a dataset collected from a consultancy activity carried out by DAS Ltd for a UK civil nuclear plant.
The data we rely on consists of 30,664 readings taken at 3 hour intervals of 14 numeric variables, amounting to 10 years of data from 2009 to 2019.
The 3 hour intervals between readings means that $\omega = 8$ , for example, corresponds to 24 hours of warning.
Over this period, 10 recorded tube leaks occurred.
As 4 of them occurred shortly after another tube leak, and, therefore, are not independent of it, the data corresponding to such leaks is discarded.
This leads us to a total of 6 fouling events.

We evaluate 11 classification algorithms used for solving Problem~\eqref{prob:3}, selected among the most used algorithms in the literature for classification tasks:
Support Vector Machine with a radial-basis-function kernel (SVM),
Random Forest (RF),
Balanced Random Forest (BRF),
Multilayer Perceptron (MLP),
Logistic Regression (LR),
AdaBoost (AB),
$k$-Nearest Neighbours (KNN),
Decision Tree (DT),
Gaussian Naive Bayes (GNB),
Quadratic Discriminant Analysis (QDA),
and Gradient Boosting (GB).
For each of them, we rely on the corresponding Python implementation available in the Python package {\tt scikit-learn}~\cite{sklearn_api,scikit-learn}.
Notice that, due to the potential need to include patterns containing an event in the training phase while excluding them from testing, we cannot use the {\tt Scikit-learn} function {\tt cross\_val\_score} for $k$-fold CV but, rather, we employ the methods presented in Section~\ref{Solving the problem}.

The only preprocessing that we apply to the data is a simple rescaling, by which we restrict each variable to a range between 0 and 1.
In our experiments, a scaler is fit to the training set and then used to scale the data from the training and testing sets.
This is implemented using the function {\tt MinMaxScaler} from {\tt Scikit-learn}.

All experiments are run using a consumer-grade personal computer, equipped with a quad-core i5 CPU and 8~GB of RAM.

\subsection{Experiment 1: Block sampling vs partitioning}

In this experiment, we compare the standard method of simply partitioning the data set into folds to our proposed block sampling method for creating folds.
In order to quantify the advantages of selecting folds using the block sampling method proposed in Section~\ref{Solving the problem}, we begin by adopting the standard approach of partitioning the entire data set into $k$ folds, where $k$ is equal to the number of events.
We go on to perform $k$-fold CV by iteratively selecting one fold for testing and using the remaining folds for training.
At each iteration of this $k$-fold CV, we will train the models using the entirety of the $k$-1 folds reserved for training.
We will also use our block-sampling method to create a block for each event contained within these $k$-1 training folds.
We will compare the average accuracy on the entire testing fold over each CV iteration of models trained on the entirety of the $k$-1 training folds with those trained on only our sampled blocks.
As an indication of which method leads to more accurate classifiers, we adopt the average testing accuracy of the classifiers trained on the entire folds compared to that of the classifiers trained on the sampled blocks.

\begin{figure}[!ht]
  \centering
  \includegraphics[width=14cm]{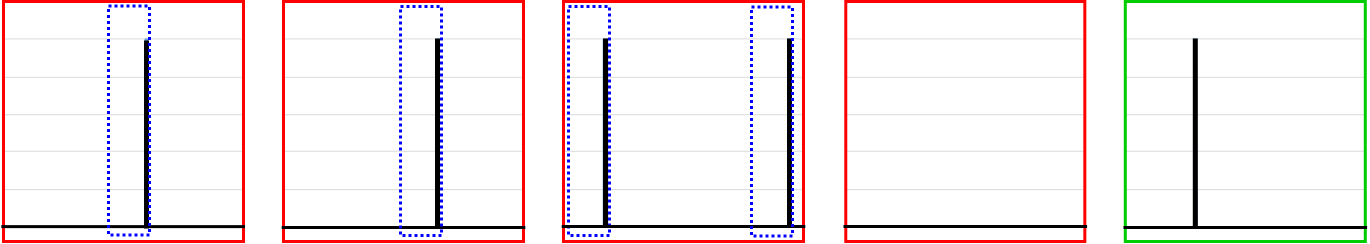}
  \caption{Foaming events, shown in black, partitioned into training and testing folds, shown in red and green, respectively.
    Blocks sampled from the training folds are shown in blue.}\label{Block_vs_Fold}
\end{figure}

Figure~\ref{Block_vs_Fold} gives an example of one iteration of this process on the AD data.
As there are 5 foaming events, the data is partitioned chronologically into 5 folds.
One of these folds is selected for testing and the remaining 4 are assigned for training.
We train the 11 classifiers on the entirety of the 4 training folds and then sample blocks from the training folds and train the 11 classifiers on these blocks.
The classifiers trained on the sampled blocks and the classifiers trained on the entire folds are then tested on the entire testing fold, and their results are compared.
The process is repeated 5 times.

As stated in Section~\ref{Solving the problem}, due to the infrequent and irregular occurrence of events, it is likely that some of the partitioned folds in this experiment will contain no events.
An example of this can be clearly seen in Figure~\ref{Block_vs_Fold}.
The inclusion of iterations in which these folds are used for testing in the calculation of the average testing accuracy would introduce a bias towards models which correctly predict that the averse event will not occur, regardless of how well they would predict that it will.
For this reason, we do not include iterations where we test folds which contain no events in the calculation of the average testing accuracy.

Through preliminary experiments, we have found that, for both applications, the inclusion of 4 lags in each data point (i.e., the adoption of $\tau = 4$) resulted in stronger classifiers.
In this initial experiment, we set $\omega$ equal to 24 and 8 for the foaming in AD and tube leakage in NPP problems respectively, corresponding to a warning of 24 hours.
When sampling using the method shown in Section~\ref{Solving the problem}, we sample 1000-hour blocks from the AD data set and 750-hour blocks from the NPP data set.
This corresponds to slightly over a month of runtime in each block.
As the AD data is recorded at hourly intervals and the NPP data is recorded at 3 hourly intervals, the sampled blocks contain 1000 and 250 patterns, respectively.

While patterns occurring within the event itself are always excluded from the testing set, we will experiment with including such patterns in the training set using the methodology laid out in Section \ref{Solving the problem}.
In Tables~\ref{Fold_Technique_Comparison_Results_AD} and \ref{Fold_Technique_Comparison_Results_NPP}, we denote models trained on data excluding the patterns occurring within the event with EE (event excluded) and those trained on data including the patterns occurring within the event with EI (event included).

\begin{table}[h!]
\small
\caption{Foaming in AD: Comparison of fold sampling techniques}
\centering
\setlength{\tabcolsep}{3pt}
\begin{tabular}{rl|| c c c c c c c c c c c | c}
 \hline
 \hline
 \multicolumn{2}{c}{Fold sampling} & \multicolumn{11}{c}{Classifier} &\multirow{2}{*}{Aver.}\\
 \cline{3-13}
 \multicolumn{2}{c}{technique} & SVM & RF & BRF	& MLP	& LR	& AB &	KNN &	DT & GNB & QDA	& GB\\
 \hline
 Basic & (EE)  &0.52	&0.50	&0.68	&0.55	&0.54	&0.55	&0.50	&0.56	&0.84	&0.51	&0.51	&0.57\\
 Folds & (EI)	&0.65	&0.52	&0.67	&0.54	&0.57	&0.56	&0.50	&0.52	&0.84	&0.59	&0.52	&0.59\\
 \hline
 Sampled & (EE)&0.70	&0.56	&0.73	&0.68	&0.64	&0.58	&0.52	&0.64	&0.83	&0.54	&0.64	&0.64\\
 Folds & (EI)  &0.74	&0.61	&0.74	&0.70	&0.65	&0.64	&0.52	&0.72	&0.87	&0.63	&0.61	&{\bf 0.68}\\
 \hline
 \hline
\end{tabular}
\label{Fold_Technique_Comparison_Results_AD}
\end{table}

\begin{table}[h!]
\small
\caption{Condenser tube leakage: Comparison of fold sampling techniques}
\centering
\setlength{\tabcolsep}{3pt}
\begin{tabular}{rl|| c c c c c c c c c c c | c}
 \hline
 \hline
 \multicolumn{2}{c}{Fold sampling} & \multicolumn{11}{c}{Classifier} &\multirow{2}{*}{Aver.}\\
 \cline{3-13}
 \multicolumn{2}{c}{technique} & SVM & RF & BRF	& MLP	& LR	& AB &	KNN &	DT & GNB & QDA	& GB\\
 \hline
 Basic & (EE)  &0.50	&0.50	&0.65	&0.50	&0.46	&0.50	&0.50	&0.50	&0.50	&0.50	&0.51	&0.51\\
 Folds & (EI)	&0.50	&0.50	&0.64	&0.50	&0.46	&0.51	&0.50	&0.51	&0.50	&0.50	&0.50	&0.51\\
 \hline
 Sampled & (EE)&0.57	&0.56	&0.73	&0.57	&0.52	&0.58	&0.52	&0.49	&0.54	&0.52	&0.49	&0.55\\
 Folds & (EI)  &0.63	&0.67	&0.71	&0.60	&0.54	&0.67	&0.56	&0.50	&0.52	&0.56	&0.64	&{\bf 0.60}\\
 \hline
 \hline
\end{tabular}
\label{Fold_Technique_Comparison_Results_NPP}
\end{table}

From inspection of Tables~\ref{Fold_Technique_Comparison_Results_AD} and \ref{Fold_Technique_Comparison_Results_NPP}, we can see that, in both applications, models trained on the sampled blocks significantly outperform models trained on the entirety of the folds.
On average, models which include event patterns in their training outperform those which do not on both the AD and the NPP data.
When including event patterns in training, on average our proposed block sampling method improves the average testing accuracy of the 11 classifiers by $15.2\%$ for the AD application and by $17.6\%$ for the NPP application.

As models trained on blocks sampled from the training folds outperform those trained on the entirety of the training folds, in the following experiments we will sample these blocks from the entire data set and disregard all remaining data.
In particular, we perform $k$-fold CV where each fold is a block containing one event.
We then evaluate the performances of the classifiers with $\omega \in \{12, 24, 36, 48, 96\}$ for the AD data set and $\omega \in \{4, 8, 12, 16, 32\}$ for the NPP data set.
These ranges of values for $\omega$ have been chosen so that the warning provided by models in each application is 12, 24, 36, 48, 96 hours.
Choosing the most appropriate value of $\omega$ depends on, firstly, how much warning we require to perform maintenance to prevent the adverse event from occurring and, secondly, on which value of $\omega$ yields the most accurate predictions.

Table~\ref{AD basic classifiers} shows the average balanced accuracy score for the 11 classifiers over a 5-fold CV for a range of values of $\omega$ for the AD application.
In this 5-fold CV, the blocks sampled from the AD data are used as folds.
The GNB classifier outperformed all others for all values of $\omega$, with CV accuracies ranging from $0.671$ to $0.875$.
The results suggest that the three classifiers achieving the best performance for predicting foaming in AD for a range of values of $\omega$ are SVM, BRF, and GNB.

\begin{table}[h!]
\small
\caption{Foaming in AD: 5-fold CV for classifier comparison}
\centering
\setlength{\tabcolsep}{3pt}
\begin{tabular}{l c c c c c c c c c c}
 \hline
 \hline
 $\omega\quad\quad$ & SVM	& BRF	& MLP	& LR	& AB &	KNN &	DT & GNB & QDA	& GB\\
 \hline
 12     &0.843	&0.757	&0.768	&0.688	&0.742	&0.541	&0.753	&{\bf 0.875}	&0.627	&0.747\\
 24     &0.759	&0.775	&0.695	&0.658	&0.715	&0.536	&0.733	&{\bf 0.873}	&0.596	&0.713\\
 36     &0.751	&0.743	&0.710	&0.686	&0.665	&0.560	&0.666	&{\bf 0.784}	&0.620	&0.674\\
 48     &0.734	&0.722	&0.640	&0.667	&0.590	&0.555	&0.599	&{\bf 0.746}	&0.605	&0.620\\
 96     &0.613    &0.628	&0.635	&0.591	&0.539	&0.591  &0.588	&{\bf 0.671}	&0.622	&0.579\\
 \hline
 \hline
\end{tabular}
\label{AD basic classifiers}
\end{table}

Table~\ref{CF basic classifiers} shows the average balanced accuracy score of 11 classifiers over a 6-fold CV for a range of values of $\omega$ for the NPP application.
In this 6-fold CV, the blocks sampled from the condenser data are used as folds.
The results suggest the three classifiers achieving the best performance for predicting condenser tube leaks for a range of values of $\omega$ are SVM, AB, and GNB, with best CV accuracies of $0.634$, $0.644$, and $0.609$ respectively.
In particular, AB performs best for small values of $\omega$, while GNB performs best for large values of $\omega$, and SVM outperforms both for $\omega = 12$.

\begin{table}[h!]
\small
\caption{Condenser tube leakage: 6-fold CV for classifier comparison}
\centering
\setlength{\tabcolsep}{3pt}
\begin{tabular}{l c c c c c c c c c c}
 \hline
 \hline
 $\omega\quad\quad$ & SVM	& BRF	& MLP	& LR	& AB &	KNN &	DT & GNB & QDA	& GB\\
 \hline
 4     &0.575	&0.603	&0.577	&0.535	&{\bf 0.608}	&0.501	&0.390	&0.577	&0.500	&0.502\\
 8     &0.633	&0.549	&0.599	&0.577	&{\bf 0.644}	&0.523	&0.455	&0.520	&0.500	&0.564\\
 12    &{\bf 0.634}	&0.545	&0.572	&0.534	&0.578	&0.517	&0.441	&0.585	&0.500	&0.547\\
 16    &0.573	&0.528	&0.552	&0.545	&0.510	&0.503	&0.515	&{\bf 0.609}	&0.520	&0.490\\
 32    &0.560	&0.537	&0.545	&0.555	&0.507	&0.520	&0.555	&{\bf 0.597}	&0.553	&0.499\\
 \hline
 \hline
\end{tabular}
\label{CF basic classifiers}
\end{table}

This experiment has shown the importance of testing a variety of classification algorithms as there is significant variety in classifier performance across different values of $\omega$ in both applications.
In the following experiments, we will attempt to improve the performance of the three best performing classifiers for each application by way of sampling and hyperparameter tuning.

\subsection{Experiment 2: Undersampling and oversampling}

As we mentioned in the previous section, we adopt under and oversampling techniques to mitigate the negative effects that class imbalance may have on the classifiers.
In the next set of experiments, we test the effects of under and oversampling on the three best-performing classifiers for each application by under or oversampling the training sets before training the classifiers and testing them on the unsampled testing sets.
As explained in the previous section, when undersampling we sample, with replacement, a set from our majority class of the same cardinality as our minority class, whereas, when oversampling, we sample, with replacement, a set from the minority class that has the same cardinality as the majority class.
This is implemented using the functions {\tt RandomUnderSampeler} and {\tt RandomOverSampler} from the Python package {\tt Imbalanced-learn}.
In Tables~\ref{AD sampled classifiers} and~\ref{CF sampled classifiers}, we use NS to denote that a classifier was trained without the use of any sampling techniques, and US or OS denote that undersampling or oversampling, respectively, was used to balance the training set.

Table~\ref{AD sampled classifiers} show the results of 5-fold CV on the blocks sampled from the AD data comparing the results of using different sampling techniques to improve the performance of the classifiers which performed best on the AD data for a range of $\omega$ values under consideration.
As expected, sampling techniques generally do not improve the performance of the BRF classifier.
This is due to the fact that BRF is a variation of the more standard random forest (RF) classifier which already integrates sampling techniques.
The GNB classifiers only sees minor increases in performance for all values of~$\omega$ when the training set is under or over sampled.
SVM classifiers show a significant average increase in accuracy of $6.7\%$ and $3.4\%$ form under and over sampling techniques, respectively.

\begin{table}[h!]
  \centering
  \caption{Foaming in AD: 5-fold CV sampling technique comparison}
  \setlength{\tabcolsep}{1pt}
  \begin{tabular}{l c c c c c c c c c c c c c c c c c c c cccc}
    \hline
    \hline
    $\omega\quad\quad$ & \multicolumn{5}{c}{SVM} & & & & & \multicolumn{5}{c}{BRF} & & & & & \multicolumn{5}{c}{GNB}\\
    \hline
                & NS & & OS & & US &     &  & & 	& NS & & OS & & US &    & &  &      &NS & & OS & & US \\
    \hline
    12  &0.843	& &0.844	  &    &{\bf 0.866}&  & &   &    &{\bf 0.757}&	&0.742 &	&0.754	&   & & &    &0.875	 &    &{\bf 0.880}&	&0.876\\
    24  &0.759	& &{\bf 0.854} & &0.821	 & &  &  &   &{\bf 0.775}	& &0.629&	&0.772	 &  & &  &   &0.873	  &   &{\bf 0.886}&	&0.876\\
    36  &0.751	& &{\bf 0.839}  & &0.829	& & &   &    &{\bf 0.743} &	&0.634&	&0.732	& &  &    &  &0.784	 &    &{\bf 0.813}&	&0.811\\
    48  &0.734	& &0.776	    & &{\bf 0.791} & &	&  &  &{\bf 0.722}& &0.606&	&0.693	& &   &  &   &0.746	  &   &{\bf 0.761}&	&0.756\\
    96  &0.613	& &0.536	    &  &{\bf 0.641}	&  & & & &0.628	    &    &0.593&	&{\bf 0.633}& & & &	&{\bf 0.671}&  &0.670	    &    &0.663\\

    \hline
    \hline
  \end{tabular}
  \label{AD sampled classifiers}
\end{table}


Table~\ref{CF sampled classifiers} shows the results of 6-fold CV on the blocks sampled from the NPP data comparing the results of using different sampling techniques to improve the performance of the classifiers which performed best on the NPP data for a range of $\omega$ values.
SVM and GNB classifiers both benefit as a result of oversampling, with an average increase in accuracy, across all tested values of $\omega$, of $10.0\%$ and $5.5\%$ respectively.
The use of sampling techniques makes the performance of the AB classifiers quite unreliable and, upon rerunning the 6-fold CV, the results from this classifier varied considerably.
For the majority of values of $\omega$, sampling techniques resulted in little to no improvement in the performance of the AB classifiers.

\begin{table}[h!]
  \centering
   \caption{Condenser tube leakage: 6-fold CV sampling technique comparison
    }
  \setlength{\tabcolsep}{1pt}
  \begin{tabular}{l c c c c c c c c c c c c c c c c c c c cccc}
    \hline
    \hline
    $\omega\quad\quad$ & \multicolumn{5}{c}{SVM} & & & & &  \multicolumn{5}{c}{AB} & & & & & \multicolumn{5}{c}{GNB}\\
    \hline
                & NS & & OS & & US &  & & & & NS & & OS & & US &  & & & & NS& & OS & & US \\
    \hline
    4   &0.575	& &{\bf 0.762}	& &0.532	& & & & & {\bf 0.608}	& &0.556	& &0.605	& & & & & 0.577	& &{\bf 0.652}	& &0.601\\
    8   &0.633	& &{\bf 0.674}	& &0.607	& & & & & 0.644	& &{\bf 0.652}	& &0.605	& & & & & 0.520	& &{\bf 0.570}	& &0.485\\
    12  &0.634	& &{\bf 0.644}	& &0.580	& & & & & {\bf 0.578}	& &0.566	& &0.550	& & & & & 0.585	& &{\bf 0.594}	& &0.568\\
    16  &0.573	& &{\bf 0.611}	& &0.535	& & & & & 0.510	& &{\bf 0.554}	& &0.512	& & & & & 0.609	& &{\bf 0.615}	& &0.597\\
    32  &0.560	& &{\bf 0.577}	& &0.554	& & & & & 0.507	& &{\bf 0.588}	& &0.497	& & & & & 0.597	& &{\bf 0.611}	& &0.599\\

    \hline
    \hline
  \end{tabular}
  \label{CF sampled classifiers}
\end{table}


This set of experiments shows that under and over sampling can lead to an often significant increase in classifier performance in both applications, with oversampling typically outperforming undersampling.
This is likely due to the fact that, with smaller values of $\omega$, the number of data patterns in the minority class is quite low, so that undersampling leads to a very small data set on which to build classifiers, which is insufficient to achieve a high accuracy.

\subsection{Experiment 3: Hyperparameter tuning}

In the previous experiments, we adopted default hyperparameter from scikit-Learn \cite{scikit-learn}.
In contrast, in this subsection, we use gridsearch to tune the hyperparameters of the three best-performing classification algorithms that we identified for each application, using a grid of roughly 100 combinations of hyperparameter values.
Due to, as explained in the previous section, the need to include, in some configurations, event patterns in the training of the models but to exclude them from testing, it is not possible to use the {\tt Scikit-learn} function {\tt GridSearchCV} for hyperparameter tuning.
Instead, we use the method laid out in Algorithm~\ref{Gridsearch}.

For the SVM classifier, we search over the values of two hyperparameters, the regularisation parameter ($\mathbf{C}$) and the kernel coefficient ($\gamma$).
We test values $10^{n}$ with $n \in \{-6,...,3\}$ for $\mathbf{C}$ and values $\frac{1}{d \, (\tau+1)}$ for $\gamma$, where $d \, (\tau+1)$ is the number of features of our data and $10^{n}$ with $n \in \{-6,...,2\}$.

For the GNB classifier, we only tune the variance smoothing hyperparameter by searching over a logarithmic range of 100 values from $10^{-15}$ to $10^{0}$.

For the BRF classifier, we tune three hyperparameters: the number of trees in the forest ($n$-estimators), the number of features to consider when looking for the best split at each node of the decision trees (max-features), and the sampling strategy used to balance the data for each tree.
We test values of $\{10,$ $25,$ $50,$ $100,$ $200,$ $300,$ $400,$ $500,$ $750,$ $1000\}$ for $n$-estimators, and of $\{2,4,8,16,32\}$ for max-features, and we test under and over sampling for balancing the data for each tree.

For the AB classifier, we test values of $\{10, 20, 30, 40, 50, 75, 100, 150, 200$, $500, 1000, 2000\}$ for the maximum number of estimators and learning rates of $10^{n}$ with $n \in \{-7,...,0\}$.

Table~\ref{AD tuned classifiers} compares the performances of the three best performing models on the AD data with tuned hyperparameters and with default hyperparameter values for a range of values of $\omega$.
Classifiers with tuned hyperparameters are marked with an asterisk.
Hyperparameter tuning improves the performance of the GNB classifiers for all values of $\omega$ and of the SVM classifiers for all values of $\omega$ but $12$.
GNB and SVM classifiers with tuned hyperparameters see an average increase in accuracy over those with preselected hyperparameters values, across all tested values of $\omega$, of $2.1\%$ and $5.0\%$ respectively.
The performance of the BRF classifiers is, in general, not improved by hyperparameter tuning, suggesting that overfitting (of the training set) occurred during the hyperparameter tuning phase.

\begin{table}[h!]
  \centering
\caption{Foaming in AD: Tuned vs not tuned classifiers}
  \begin{tabular}{l c c c c c c cccc}
    \hline
    \hline
    $\omega\quad\quad$ & SVM	& SVM*  & & & BRF	& BRF*	& & & GNB &	GNB*\\
    \hline
    4  &{\bf 0.843}	&0.830	   & & & {\bf 0.757}	&0.692	& & & 0.875	&{\bf 0.882}\\
    8  &0.759	&{\bf 0.795}	& & & {\bf 0.775}	&0.729	& & & 0.873	&{\bf 0.877}\\
    12 &0.751	&{\bf 0.806}	& & & {\bf 0.743}	&0.733	& & & 0.784	&{\bf 0.822}\\
    16 &0.734	&{\bf 0.778}	& & & 0.722	&{\bf 0.735}	& & & 0.746	&{\bf 0.776}\\
    32 &0.613	&{\bf 0.667}	& & & {\bf 0.628}	&0.609	& & & 0.671	&{\bf 0.674}\\
    \hline
    \hline
  \end{tabular}
  \label{AD tuned classifiers}
\end{table}

Table~\ref{CF tuned classifiers} compares the performances of the three best-performing classifiers on the NPP data with tuned hyperparameters and with default hyperparameter values for a range of values of $\omega$.
Classifiers with tuned hyperparameters are marked with an asterisk.
As the table indicates, for this dataset hyperparameter tuning does not consistently improve the performance and, when it does, the improvement is typically very minor.

\begin{table}[h!]
\centering
\caption{Condenser tube leakage: Tuned vs not tuned classifiers}
  \begin{tabular}{l c c c c c c cccc}
    \hline
    \hline
    $\omega\quad\quad$ & SVM	& SVM*	& & & AB	& AB*	& & & GNB &	GNB*\\
    \hline
    4  &0.575	&{\bf 0.576}	& & &{\bf 0.608}	&0.584	& & &0.577	&{\bf 0.588}\\
    8  &{\bf 0.633}	&0.573	    & & &{\bf 0.644}	&0.614	& & &0.520	&{\bf 0.587}\\
    12 &0.634	&0.634	        & & &{\bf 0.578}	&0.558	& & &{\bf 0.585}	&0.580\\
    16 &0.573	&{\bf 0.577}	& & &0.510	&{\bf 0.519}	& & &{\bf 0.609}	&0.578\\
    32 &{\bf 0.560}	&0.473	    & & &{\bf 0.507}	&0.496	& & &{\bf 0.597}	&0.522\\
    \hline
    \hline
  \end{tabular}
  \label{CF tuned classifiers}
\end{table}

To summarise, these experiments show that, while hyperparameter tuning significantly increases the performance on the AD data set, it results in little to no improvement on the NPP data set.

\subsection{Final recommendations}

We conclude this section by reviewing the different results we obtained from our 3 experiments.
We will collect the performances of the best performing classifiers and note where they use sampling schemes and hyperparameter tuning.

\begin{figure}[!ht]
  \centering
  \includegraphics[width=14cm]{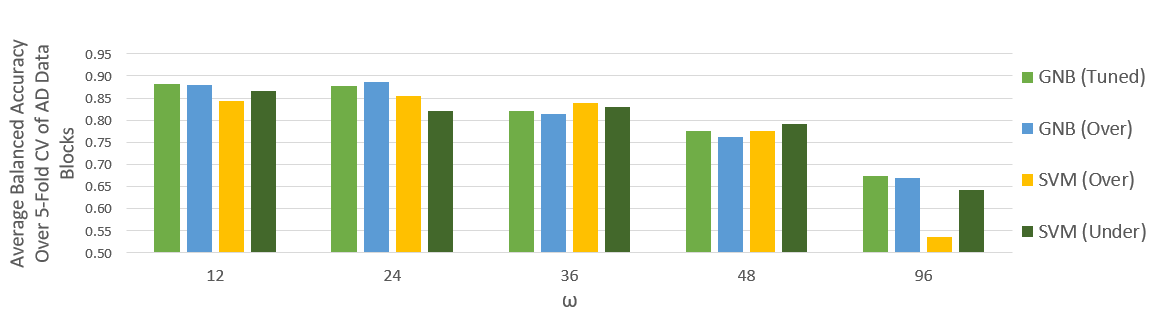}
  \caption{Graph of the average testing accuracies of the classifiers which perform best at one or more value of $\omega$ in predicting foaming in AD }\label{Foaming_Best_Classifiers}
\end{figure}

Figure~\ref{Foaming_Best_Classifiers} shows the average testing accuracy of the best classifiers for various values of~$\omega$ over 5-fold CV of the AD data.
Each classifier included in this graph achieved the highest average testing accuracy of all classifiers tested for at least one value of $\omega$.
Though sampling techniques and hyperparameter tuning both improved the performance of our best-performing classifiers (SVM and GNB), tuning hyperparameters in combination with sampling the training set did not lead to good classifiers.
In light of these results, $\omega = 24$, corresponding to 24 hours of warning, results in the most accurate models whilst giving sufficient warning to carry out necessary maintenance to prevent foaming.
For this value of $\omega$, the best performing classifier is the GNB classifier using over-sampling, which achieves an average testing accuracy of $0.886$.

\begin{figure}[!ht]
  \centering
  \includegraphics[width=14cm]{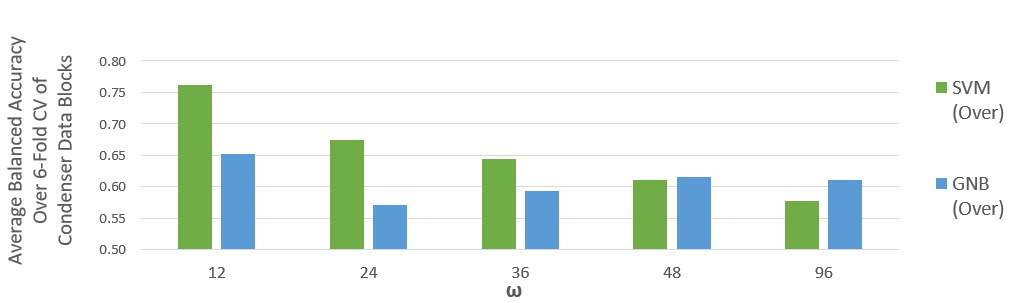}
  \caption{Graph of the average testing accuracies of the classifiers which perform best at one or more value of $\omega$ in predicting tube leaks in NNP}\label{EDF_Best_Classifiers}
\end{figure}

Figure~\ref{EDF_Best_Classifiers} shows the average balanced accuracy of the best models for various values of 6-fold CV of the NPP data blocks.
Each classifier included in this graph had the highest average testing accuracy of all classifiers tested for at least one value of $\omega$.
Unlike with the AD data set, where a range of classifiers performs well at different values of $\omega$, here we have one classifier outperforming all others for the majority of tested values of $\omega$.
In particular, the SVM classifier using oversampling outperforms all other classifiers for $\omega$ less than or equal to 36.
For values of $\omega$ larger than 36, the GNB classifier using oversampling gives the best performance however all classifiers.
For this reason, we recommend to adopt the smallest value of $\omega$ which would still allow time for preventative measures to be undertaken.

\begin{figure}[!ht]
  \centering
  \includegraphics[width=14cm]{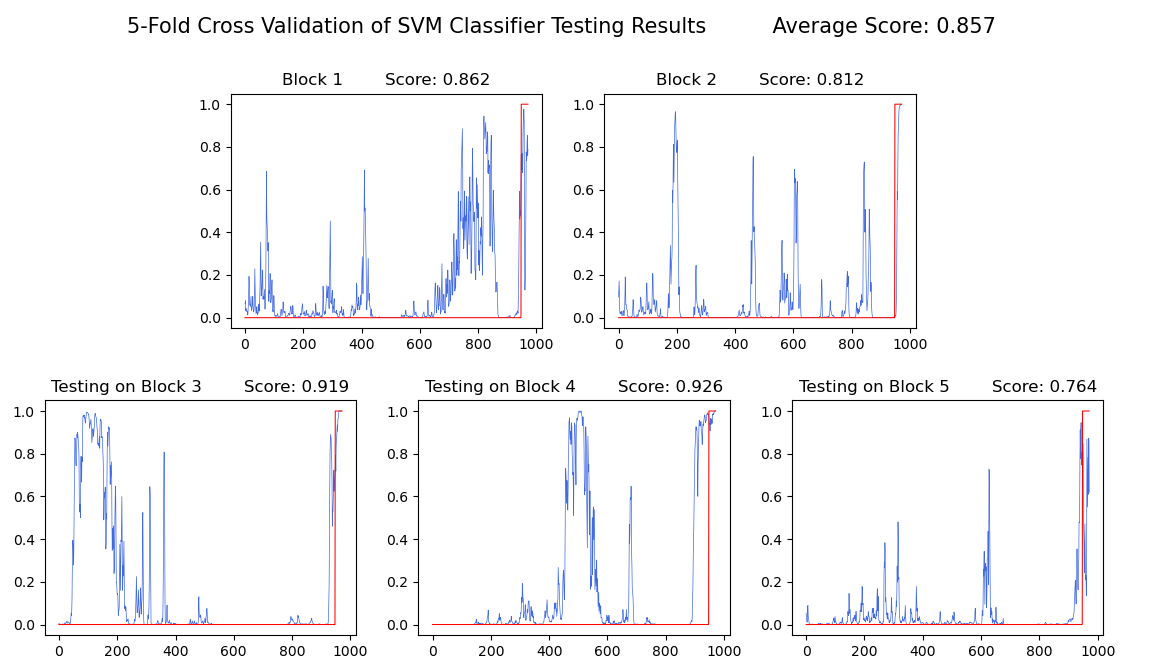}
  \caption{Example of the SVM using oversampling's performance as a potential predictive maintenance tool for predicting foaming in AD. Each plot represents the testing on one block of a model trained on the remaining 4 blocks. The model's predicted likelihood of foaming is shown in blue and the warning label is shown in red.}\label{SVM_Over_Sampled_24h_Results}
\end{figure}

Figure~\ref{SVM_Over_Sampled_24h_Results} shows the testing predictions of one of the best classifiers for preempting foaming in AD, SVM, with a 24 hour event association which had an average testing accuracy of $0.854$.\footnote{We report the results of SVM with oversampling despite the fact that, with $\omega = 24$, such a classifier is outperformed by GNB with oversampling or tuned hyperparameters.
The choice is motivated by the fact that the predicted probability outputs from the GNB classifiers are almost binary, implying that the graph of its outputs are quite difficult to interpret.}
Each plot shows a classifier trained on four of the blocks and tested on the remaining block.
The blue line shows the predicted probability of foaming and the red line shows the corresponding warning label $\tilde y_t$.
As can be seen, this classifier is able to accurately preempt each one of the foaming events given that it has been trained on the remaining four events.

The results for predicting condenser tube leaks in the NPP application could be plotted similarly.
With $\omega = 8$, corresponding to 24 hours of warning, our best classifier, SVM using oversampling, is capable of preempting 3 of the 6 tube leaks with a balanced accuracy of at least 0.75 on each block.
If it is acceptable to reduce $\omega$ to 4, corresponding to 12 hours of warning, then we can preempt 4 of the 6 tube leaks with a balanced accuracy of at least 0.75 on each block.

 \section{Conclusions}\label{Conclusion}

We have proposed a framework for formulating and solving a classification problem based on adverse event prediction with focus on two applications: predicting foaming in anaerobic digesters and predicting condenser-tube leaks in nuclear power plants.

Our experiments have revealed that our proposed framework achieves a good performance on the data sets of the two applications we considered, showing that it is possible to predict foaming in anaerobic digesters and condenser tube leaks in nuclear power production with reasonable accuracy.

In a practical industrial context, the predicted maintenance models derived with our techniques can effectively reduce the expenditure on anti-foaming agents in the case of anaerobic digestion and the cost of condenser fouling that are due to the reduced power generation in the production of nuclear energy.


%

\section*{Acknowledgement}
The work of AJD is jointly funded by Decision Analysis Services Ltd and EPSRC through the Studentship with Reference EP/R513325/1.
The work of ABZ is supported by the EPSRC grant EP/V049038/1 and the Alan Turing Institute under the EPSRC grant EP/N510129/1.

The authors would like to thank DAS Ltd for the support provided with the work summarised in this paper.
In particular, they would like to thank Jack Lewis for encouraging this research work and for the logistical and technical support in the understanding of the data sets.
They would also like to extend their gratitude to Daniel Everitt and Nicholas Barton for regular discussions that helped in refining the models. Si\^{o}n Cave's feedback on the initial draft of the paper is also gratefully acknowledged.


\bibliographystyle{plain}
\bibliography{bibliography}

%
%
%

\end{document}